\pdfoutput=1
\documentclass[11pt]{article}

\usepackage[preprint]{acl}

\usepackage{times}
\usepackage{latexsym}
\usepackage{subcaption}
\usepackage{amsmath}
\usepackage{mathtools}

\usepackage{amsfonts}
\usepackage{tabularray}
\usepackage{multirow}
\usepackage{booktabs}

\usepackage[T1]{fontenc}
\usepackage[utf8]{inputenc}

\usepackage{microtype}

\usepackage{inconsolata}

\usepackage{graphicx}

\DeclareMathAlphabet{\mathcal}{OMS}{cmsy}{m}{n}

\title{Mind Your Theory: Theory of Mind Goes Deeper Than Reasoning}
\author{
\textbf{Eitan Wagner$^*$\textsuperscript{1}} \quad
\textbf{Nitay Alon$^*$\textsuperscript{1,2}} \quad
\textbf{Joseph M. Barnby\textsuperscript{3,4}} \quad
\textbf{Omri Abend\textsuperscript{1}}
\\
\\
\textsuperscript{1} Hebrew University of Jerusalem \quad
\textsuperscript{2} MPI for Biological Cybernetics \\
\textsuperscript{3} Centre for AI and Machine Learning, AU \quad
\textsuperscript{4} IoPPN, King's College London \\
\texttt{\{eitan.wagner, nitay.alon\}@mail.huji.ac.il}
}

\begin{document}

\maketitle
\def\thefootnote{*}\footnotetext{Equal contribution.}\def\thefootnote{\arabic{footnote}}

\begin{abstract}
Theory of Mind (ToM) capabilities in LLMs have recently become a central object of investigation, sparking debates and discussions. In this position paper, we explore many lines of work in different communities in AI and cognitive science. Inspired by cognitive work, we view ToM tasks as a two-step process: (I) first, determining whether and how to invoke ToM, which includes setting the appropriate Depth of Mentalizing (DoM); and (II) second, applying correct inference given the appropriate DoM. 
We identify that many works about ToM in LLMs, such as benchmarks and add-on modules, tend to unjustly overlook the first step and focus exclusively on the second one, which can be framed as a logic-reasoning task. We support our distinction with empirical evidence about the difficulty of the different steps in existing benchmarks.
We conclude with suggestions for improved evaluation of ToM capabilities, inspired by dynamic environments used in cognitive tasks in biological agents.
\end{abstract}

\section{Introduction}
The ability of Large Language Models (LLMs) to efficiently integrate social information is essential for ensuring AI trust and safety, as well as for reasoning about situations involving multiple agents. This capability, also known as mentalizing, or Theory of Mind (ToM) \cite{premack1978does}, relies on inferring and representing otheragents' beliefs, desires, and intentions.

LLMs have been recently regarded as general-purpose reasoning models \cite{brown2020languagemodelsfewshotlearners}, and their ToM capabilities have come under scrutiny, particularly as to whether social cognition can emerge purely from associative principles \cite{sap-etal-2022-neural}. Notably, \citet{kosinski2023theory} claimed that ToM capabilities emerge in post-GPT3 models, like ToM in children \cite{astington1995theory}. This claim sparked debates on the correct interpretation of these results \cite{ullman_large_2023, pi2024dissectingullmanvariationsscalpel,strachan2024testing}.

\begin{figure}[t]
\centering
\includegraphics[width=0.49\textwidth]{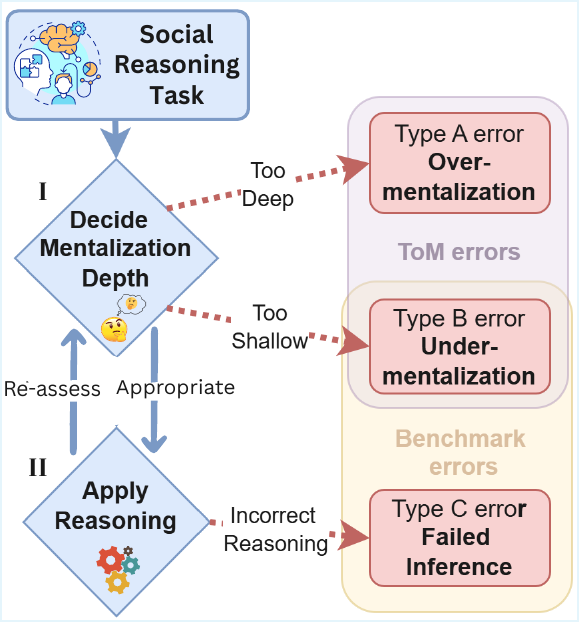}
\caption{Overview of ToM error types. 
Existing benchmarks address cases of clear ToM 
intention and deal mainly with step II (and Type C errors), overlooking the possibility of Type B errors.
In contrast, the decision to invoke ToM (step I) should address Type A and Type B errors.}
\label{fig: flow}
\end{figure}

Existing benchmarks for ToM primarily concentrate on whether agents have correct beliefs about others \cite{kosiniski_llm_tom_bench_mark_review}. However, within the context of social interaction, ToM capabilities involve two steps \cite{schaafsma2015deconstructing, leslie2004core}: (1) determining the depth of mentalization \cite[DOM;][]{barnby2023formalising} to use, which specifically includes the decision of whether to model the agents separately (self-other divide); and (2) applying correct inference for each agent's mental state, given the depth. Failure in any step leads to an incorrect conclusion (see Figure \ref{fig: flow}). 
We argue that very little attention is paid to whether LLMs correctly invoke ToM (when needed). Determining this is critical: (1) it ignores a critical theoretical component of ToM that has been well-defined in humans; (2) it confounds ToM errors with poor or fuzzy logic \cite{pi2024dissectingullmanvariationsscalpel}, and (3) the act of mentalizing has a price, both in resources and performance \cite{rilling2004neural, keysar2003limits, devaine2014theory}, and so more precisely understanding moment-to-moment ToM in LLMs may improve compute efficiency at equivalent parameter sizes. Of note, we do not focus on the evolutionary pressures and resource-rational necessities of social representation -- the \emph{why} of ToM -- and merely focus on the \emph{how}. %Why ToM exists and may be important for biological (and potentially synthetic) agents is outside the scope of this paper.

In this paper, we first discuss contemporary approaches to ToM in AI research at a high level. This includes LLM benchmarks, ToM add-ons for LLMs, and agent-focused ToM models. We identify that the focal point of current LLM and AI
research is logical inference \cite{verbrugge2008learning}, corresponding to step II (and Type C errors), which is based on the determination of DoM in the first stage. We experiment with two common ToM benchmarks to demonstrate this conclusion.
We then turn to the cognitive science literature and identify differences in the definition of ToM behind biological and synthetic agents, as the former focuses also on step I -- the act of mentalizing, and not only on correct inference. We finally identify ways to improve ToM research in AI, inspired by work in biological agents.

Our contributions are:
(1) We explore and discuss ways of addressing ToM in various fields.
(2) We formally describe ToM tasks as a two-step task, inspired by cognitive science.
(3) We identify a lacuna in existing LLM works which tend to overlook the first step of mentalizing and focus instead on logic problems.
% two key components of ToM - intentionality and adaptability.
(4) We provide a roadmap for appropriate evaluation of ToM invocation inspired by biological agents.

\section{ToM in AI research}

Here we analyze
%\oa{in what sense?} 
the various approaches and challenges used to model and assess ToM in the AI communities. We begin with a review of existing AI methods before discussing LLM-focused approaches. 

\subsection{Formal Architectures of ToM}
%\oa{what is the context of this subset? how does it related to the rest of the paper?}

Inspired by Reinforcement Learning (RL), ToM has been modeled as an inferential process in which an agent makes inferences about other agents \cite{baker2011bayesian}. This approach is similar to Inverse-RL \cite{ng2000algorithms, jara-ettinger_theory_2019}, where an inferring agent learns an association between actions and unobserved variables, such as goals or future actions. 

\begin{figure}[t]
\centering
\includegraphics[width=0.49\textwidth]{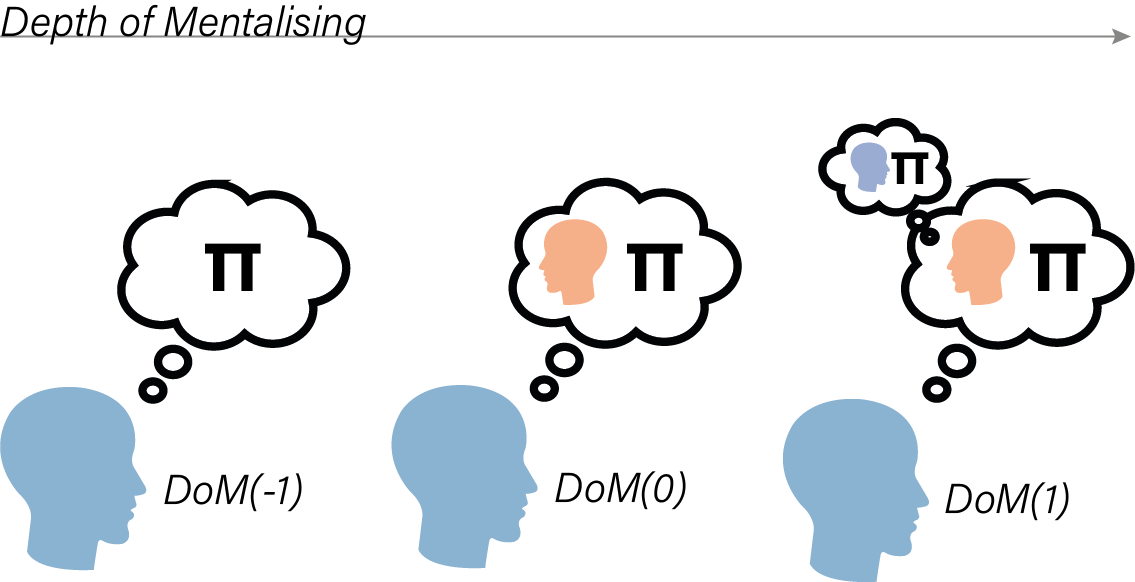}
\caption{Hierarchical Dynamic Theory of Mind (ToM). Depth of Mentalizing (DoM) indicates how much an agent (blue) recursively nests beliefs about their partner (orange). At DoM(-1) the agent only considers their own policy ($\pi$). DoM(0) beliefs consider $\pi$ and a partner's $\pi$. DoM(1) beliefs consider DoM(0) and the beliefs of the partner about the agent's $\pi$.}
\label{fig: hi-tom}
\end{figure}

The main characteristic of ToM is the depth of recursion, also known as Depth of Mentalizing (DoM). 
In observational tasks (e.g., SA tasks), the inferring agent is typically modeled as level-zero Depth of Mentalizing (DoM) -- DoM$(0)$ \cite{barnby2023formalising}. This model does not allow for recursive beliefs (``I think that you think that I think''). Instead, it is limited to shallower mentalizing (``I think that'') - with an absence of expected knowledge of the observer (see Figure \ref{fig: hi-tom}). 

Higher DoM levels (like the one presented in \citet{wu-etal-2023-hi}) are modeled using Interactive-POMDPs \cite[IPOMDP;][]{gmytrasiewicz_framework_2005}, which model other agents as part of the world model. We refer the reader to \citet{alon_disinformation} for a description of the model and its applications in multi-agent environments. 

Some attempts have been made to integrate recursive formulations into LLMs. However, due to the black-box nature of LLMs, this is difficult to achieve \cite{zhang2023building, hu2023languagemodelsagentmodels}.

\subsection{ToM Benchmarking} \label{sec:benchmarking}
% ToM works on LLMs typically use the definition from ...
% Then they ask whether LLMs possess this ability.
% \ew{I can make this section shorter if necessary}
Evaluation of LLMs on ToM benchmarks is an active line of research. A dominant component in these tests is variations of the Sally-Anne (SA) test \citep{ wimmer1983beliefs, baron1985does}. In the SA test, participants are tested on their ability to identify false beliefs, by recognizing that Anne has moved a ball while Sally, who is not exposed to this event, maintains the belief that the ball did not move. These correspond to observational inference tasks \cite{miura2021unifying}.

Variations in the SA task have been used as a gold standard benchmark for ToM in LLMs \cite{grant2017can, nematzadeh-etal-2018-evaluating, gandhi2023understandingsocialreasoninglanguage}.
Benchmarks like ToMi \cite{le-etal-2019-revisiting} and OpenToM \cite{xu-etal-2024-opentom}, improve previous benchmarks with less predictable generation protocols and more natural settings.
Rather than vignette-based tasks, some benchmarks use a language-based conversational setting to assess ToM reasoning, such as Mindcraft \cite{bara-etal-2021-mindcraft}, FANToM \cite{kim-etal-2023-fantom}, and NegotiationToM \cite{chan-etal-2024-negotiationtom}.
Some benchmarks extend the SA task to more complex objectives. This includes higher-order ToM, as in HiToM \cite{wu-etal-2023-hi}; additional perception inference, as in Percept-ToMi and Percept-FANToM \cite{jung-etal-2024-perceptions}; and epistemic logic conclusions \cite{sileo-lernould-2023-mindgames}.
Additional extensions derive questions from common ground annotations, as in CommonToM \citep{soubki-etal-2024-views} or from other modalities, as in MMToM-QA \citep{jin-etal-2024-mmtom}.

The utility of these tasks in assessing ToM in LLMs is still a matter of debate. GPT3 showed difficulties engaging with ToMi \cite{sap-etal-2022-neural}, whereas more complex models have been argued to show more success \cite{kosinski2023theory, gandhi2023understandingsocialreasoninglanguage}. Nevertheless, slight alterations to task structure or details can alter the performance \cite{ullman_large_2023, shapira-etal-2024-clever, ma-etal-2023-tomchallenges, chen-etal-2024-tombench, nickel2024probingrobustnesstheorymind}. The theoretical implications of these effects are therefore still under debate
\cite{pi2024dissectingullmanvariationsscalpel}. Other criticisms of current benchmarks are due to the lack of transparency of training sets, hindering the ability to compare LLMs with human performance \cite{ma-etal-2023-towards-holistic, shapira-etal-2024-clever}.

In particular, all these benchmarks are static in the sense that they deal with pure observation rather than the evolution of the moment-to-moment model during interaction \cite{ma-etal-2023-towards-holistic}. They therefore only distantly resemble the application of ToM capabilities in real-world settings. Additionally, these benchmarks assume that there is always a single correct answer, for which all the relevant information is given, rather than a range of possible beliefs and behaviors. 

\subsection{ToM Add-ons}

Many methods were proposed for improving ToM capabilities in existing architectures. We describe these methods as \textit{ToM Add-ons}. These components attempt to complement LLMs with external modules, assuming it is necessary.
% assume the LLM in itself is insufficient.

%\oa{are these examples of add ons?} 
There are various types of suggested add-ons. Task-specific prompting techniques have shown improved scores on classic ToM tasks \cite{tan2024phantompersonabasedpromptingeffect, jung-etal-2024-perceptions, huang-etal-2024-notion, wilf-etal-2024-think}. Explicit symbolic modules can also be updated based on input and improve accuracy \cite{qiu-etal-2024-minddial, hou-etal-2024-timetom}. Others add post-model decoding methods \cite{sclar-etal-2023-minding} or symbolic planners \cite{jin-etal-2024-mmtom} to improve performance.

\section{What Can We Learn From Biological Agents?}

Cognitive science, Economics, and Neuroscience have explored ToM in humans for decades. Insights gained over the years can inform the development of architectures in LLMs and AI in general, but regrettably, this literature is seldom referenced in NLP. Particularly, in these fields researchers explored adaptability --- changes in the DoM level of an agent during an interaction in response to feedback \citep{grosskopf2007rational} and intentionality --- invoking ToM when needed and avoiding it when unneeded \citep{aichhorn2006visual}.

We formalize these elements of ToM as follows: given a problem (=prompt) $X$ and a variable of interest $Y$ (e.g., the correct location of the object in an SA test), our general goal is to infer the top class based on $P(Y|X)$.
Inference consists of two steps (see Figure \ref{fig: flow}): (I) first the meta-decision regarding the DoM, and then (II) the reasoning process of attributing correct beliefs to agents, given the DoM.
In terms of $do$-calculus \cite{pearl2012calculus}, the first step corresponds to $do(\text{DoM(k)})$ given $X$, and the second step corresponds to $P(Y|do(\text{DoM}(k)), X)$.

More generally, the first step corresponds to the act of ``theorizing'' about the agents' belief structure, and the second step corresponds to the application of this theory. The theory can be reassessed when new information is available, forming a dynamic theory. 

\subsection{Invoking DoM}

Several theoretical and empirical works identify the common principles of social interaction. Biological agents performing cooperative tasks can afford very shallow mentalizing \cite{devaine2014theory}, and instead rely on a common policy (e.g., social norms or prosocial goals). When competition between agents is present, hierarchical mentalizing is beneficial to avoid duplicitous action \cite{alon_disinformation}. However, moving from a shallow to a hierarchical level of mentalizing is more expensive, requiring additional time and energy to compute \citep{sweller1988cognitive, bossaerts2017computational}. This limits the ability to perform ToM-demanding tasks for a long time \citep{rilling2004neural}, and can lead to failures in social reasoning when ToM is needed but jeopardized by disconnected typical neural circuits \cite{barnby2022increased}. 
While LLMs do not have the same energy constraints as humans, failures to appropriately capture the degree to which LLMs can move into competitive modes of mentalizing may hinder human-AI alignment.

Even with properly invoked ToM reasoning, an agent may still suffer from maladaptive DoM. For example, remaining fixed in a shallow DoM may miss malicious intent from other agents \cite{sarkadi2019modelling}. On the other hand, invoking hierarchical DoM during a cooperative problem can induce false beliefs of harm \cite{alon2024mal}. Moreover, as described above, when interacting with agents with varying DoM levels \citep{camerer2003cognitive}, a ``correct'' DoM level may be ill-defined. Arguably, modeling DoM involves estimating a distribution, as opposed to question-answering scenarios which involve predicting a single response \citep{wagner2025languagemodelprobabilitiesrepresent}.

\subsection{Revisiting Existing Work}

\paragraph{What do the ToM benchmarks benchmark?}
With current vignette-based reasoning tasks, the necessity of ToM is made obvious. Thus, current benchmarks practically address step II of our formulation, thus precluding the need to use ToM or even reason about self vs. other distinctions.
Even as challenging benchmarks are introduced, the focus is still on non-interactive presentations with templatized questions, in which the exact required DoM level, $k$, is rather obvious from the question. 
The problems these benchmarks address are therefore logical ones, which, while challenging, ignore crucial aspects of ToM.

We experimentally support this claim on two challenging false-belief benchmarks, HiToM \cite{wu-etal-2023-hi} and FANToM \cite{kim-etal-2023-fantom}. In addition to the belief-related questions arising from the described scenarios, we add a task of predicting the required ToM order for each question. The ToM order is recorded as metadata as it guides the generation process of the scenarios, providing gold-truth labels. 
Arguably, accurately identifying the required ToM order while incorrectly answering questions about the beliefs, shows that the main challenge is not in the ToM invocation step. Our results show this is indeed the case: while GPT-3.5-turbo struggles even with this task, GPT-4o-mini performs well, and GPT-4o is nearly perfect. Chain-of-thought (CoT) prompting gives some improvement. See Appendix \ref{appendix:a} for full details of the experiments and results.

% \ew{we can add that the process of adding more and more papers is essentially making the reasoning part harder but ignores the core question of ToM}

We note that even if $DoM$ is not given, as long as we test the prediction for $Y|X$, we cannot determine whether performance reflects ToM capabilities. If the model is incorrect, if we do not differentiate between the steps, we will not know if the failure was due to undermentalization (Type B error) or due to failed reasoning (Type C error). If the model is correct but ToM is unnecessary, we cannot tell if the success was based on the correct ToM (Type A error) \cite{kim-etal-2023-fantom}.
Only if ToM is necessary and the model succeeded, we may conclude that ToM was used and correctly so (with the caveat of spurious correlations).

\paragraph{What do ToM add-ons add?}
Typical ToM add-on works propose methods for specific ToM tasks. Essentially, these methods assume ToM is required ($DoM(k\geq 0)$), thus focusing on step II. The decision about the required DoM level (step I) is not directly addressed.

\paragraph{What do formal models model?}
Typical works on ToM propose models that inherently assume a fixed DoM$(k), k \geq 0$. While these models are beneficial for learning at this level, the fixed, immutable DoM level assumption (with some exceptions like \citet{camerer2003cognitive} and \citet{grosskopf2007rational}) limits the ability to properly model human data, where DoM levels may change. Moreover, they do not address the step of invoking ToM (step I) -- they assume that ToM is needed and invoked for inference within a given task. A more complete model that examines both steps is, to the best of our knowledge, missing from this literature. 

\paragraph{Why is this important?}
The influence of cognitive science is evidenced in many ToM benchmark papers. Many works \citep[e.g., ][]{kim-etal-2023-fantom, jin-etal-2024-mmtom} mention and adopt psychology principles that are required when designing a task to validate ToM. Additionally, the active debate about emerging ToM capabilities (\S \ref{sec:benchmarking}) is highly connected to the assumption that ToM capabilities reflect more than a specific logical task. In general, there seems to be an underlying assumption that ToM capabilities in models should cover the same tasks that humans perform by using ToM.
Many works \citep[e.g., ][]{ullman_large_2023, shapira-etal-2024-clever} question the generalization and robustness of LLM performance within a specific ToM task. Our formalization further suggests that a crucial cognitive part of ToM behavior in humans, which has its unique properties, is rarely addressed in common LLM literature.

\section{Discussion and Conclusion}

In this work, we cast ToM in LLMs as an interdisciplinary topic. We argue that the problems addressed by existing NLP benchmarks and AI works are essentially non-interactive logical problems.
While many of the benchmarks are valuable as reasoning tests and may provide insights into the performance of LLMs, they do not address the core issue of ToM in LLMs: interaction, and social adaptation.

We argue that a better understanding of ToM in LLMs requires:
(1) benchmarks that test whether DoM is correctly invoked or not, (2) evaluation for ToM in interactive settings where the model is an active agent, and (3) distributional assessment of DoM dynamics, given a social context (cooperative vs competitive). 

The requirement for social dynamism is expressed in cognitive science and computational psychology. 
Neural outcomes show distinct representations when considering non-social versus social outcomes \cite{zhang2020brain}, demanding a theoretical distinction that captures social dynamics sufficiently \cite{cushman2024computational, jara-ettinger_theory_2019, barnby2023formalising, barnby2024beyond}. We believe this distinction should also be reflected in the way we benchmark models for ToM.

\citet{wang2025rethinkingtheorymindbenchmarks}, argue in favor of dynamic, user-centric benchmarks for ToM evaluation. Our work extends this call, emphasizing the role of dynamic settings for a phase in the ToM reasoning that is commonly overlooked. 
Indeed, some previous works introduce ToM tasks for LLMs or formal models in interactive domains \cite{sclar2022symmetric,li-etal-2023-theory, ying_truong_tenenbaum_gershman_2025}. However, these benchmarks are limited in their scope as they tend to model simplified fixed worlds. We believe that a meaningful understanding of ToM capabilities should be achieved by extending works like these with additional domains and contexts. Whereas a single scenario may not fully capture the range of behavior, constructing multiple diverse ones should lead to better coverage of ToM complexities. This should include situations without full context (e.g., without explicit information about the requested), social situations (e.g., with varying levels of cooperation), and ToM behavior of other agents (e.g., heterogeneous DoM levels in the population). 

\section*{Limitations}

Our work focuses on attributed beliefs, which are a single component of ToM. Previous work has pointed out that true evaluation of ToM must include many other factors, such as emotions and desires \cite{ma-etal-2023-towards-holistic}. Despite this limitation, we note that belief attribution remains a critical component in ToM evaluation.

We also note that the ToM literature in cognitive science is vast and often conflicting. The components of ToM are actively debated; thus our recommendations must be interpreted in light of this.

\section*{Acknowledgments}
This work was supported in part by the Israel Science Foundation (grant no. 2424/21). NA is supported by the Max Planck Society and the Humboldt Foundation.

\bibliography{latex2}

\begin{thebibliography}{67}
\providecommand{\natexlab}[1]{#1}

\bibitem[{Aichhorn et~al.(2006)Aichhorn, Perner, Kronbichler, Staffen, and Ladurner}]{aichhorn2006visual}
Markus Aichhorn, Josef Perner, Martin Kronbichler, Wolfgang Staffen, and Gunther Ladurner. 2006.
\newblock Do visual perspective tasks need theory of mind?
\newblock \emph{Neuroimage}, 30(3):1059--1068.

\bibitem[{Alon et~al.(2024)Alon, Schulz, Bell, Moutoussis, Dayan, and Barnby}]{alon2024mal}
Nitay Alon, Lion Schulz, Vaughan Bell, Michael Moutoussis, Peter Dayan, and Joseph~M Barnby. 2024.
\newblock (mal) adaptive mentalizing in the cognitive hierarchy, and its link to paranoia.
\newblock \emph{Computational Psychiatry}, 8(1):159.

\bibitem[{Alon et~al.(2023)Alon, Schulz, Rosenschein, and Dayan}]{alon_disinformation}
Nitay Alon, Lion Schulz, Jeffrey~S. Rosenschein, and Peter Dayan. 2023.
\newblock \href {https://doi.org/10.1162/opmi_a_00097} {A (dis-)information theory of revealed and unrevealed preferences: Emerging deception and skepticism via theory of mind}.
\newblock \emph{Open Mind}, 7:608--624.

\bibitem[{Astington and Jenkins(1995)}]{astington1995theory}
Janet~Wilde Astington and Jennifer~M Jenkins. 1995.
\newblock Theory of mind development and social understanding.
\newblock \emph{Cognition \& Emotion}, 9(2-3):151--165.

\bibitem[{Baker et~al.(2011)Baker, Saxe, and Tenenbaum}]{baker2011bayesian}
Chris Baker, Rebecca Saxe, and Joshua Tenenbaum. 2011.
\newblock Bayesian theory of mind: Modeling joint belief-desire attribution.
\newblock In \emph{Proceedings of the annual meeting of the cognitive science society}, volume~33.

\bibitem[{Bara et~al.(2021)Bara, CH-Wang, and Chai}]{bara-etal-2021-mindcraft}
Cristian-Paul Bara, Sky CH-Wang, and Joyce Chai. 2021.
\newblock \href {https://doi.org/10.18653/v1/2021.emnlp-main.85} {{M}ind{C}raft: Theory of mind modeling for situated dialogue in collaborative tasks}.
\newblock In \emph{Proceedings of the 2021 Conference on Empirical Methods in Natural Language Processing}, pages 1112--1125, Online and Punta Cana, Dominican Republic. Association for Computational Linguistics.

\bibitem[{Barnby et~al.(2024)Barnby, Bellucci, Alon, Schilbach, Frith, and Bell}]{barnby2024beyond}
JM~Barnby, G~Bellucci, N~Alon, L~Schilbach, CD~Frith, and V~Bell. 2024.
\newblock Beyond theory of mind: A formal interoperable framework for social inference and representation.

\bibitem[{Barnby et~al.(2023)Barnby, Dayan, and Bell}]{barnby2023formalising}
Joseph~M Barnby, Peter Dayan, and Vaughan Bell. 2023.
\newblock Formalising social representation to explain psychiatric symptoms.
\newblock \emph{Trends in cognitive sciences}, 27(3):317--332.

\bibitem[{Barnby et~al.(2022)Barnby, Dean, Burgess, Kim, Teunisse, Mackenzie, Robinson, Dayan, and Richards}]{barnby2022increased}
Joseph~M Barnby, Ryan~J Dean, Henry Burgess, Jeffrey Kim, Alessa~K Teunisse, Lisa Mackenzie, Gail~A Robinson, Peter Dayan, and Linda~J Richards. 2022.
\newblock Increased persuadability and credulity in people with corpus callosum dysgenesis.
\newblock \emph{Cortex}, 155:251--263.

\bibitem[{Baron-Cohen et~al.(1985)Baron-Cohen, Leslie, and Frith}]{baron1985does}
Simon Baron-Cohen, Alan~M Leslie, and Uta Frith. 1985.
\newblock Does the autistic child have a “theory of mind”?
\newblock \emph{Cognition}, 21(1):37--46.

\bibitem[{Bossaerts and Murawski(2017)}]{bossaerts2017computational}
Peter Bossaerts and Carsten Murawski. 2017.
\newblock Computational complexity and human decision-making.
\newblock \emph{Trends in cognitive sciences}, 21(12):917--929.

\bibitem[{Brown et~al.(2020)Brown, Mann, Ryder, Subbiah, Kaplan, Dhariwal, Neelakantan, Shyam, Sastry, Askell, Agarwal, Herbert-Voss, Krueger, Henighan, Child, Ramesh, Ziegler, Wu, Winter, Hesse, Chen, Sigler, Litwin, Gray, Chess, Clark, Berner, McCandlish, Radford, Sutskever, and Amodei}]{brown2020languagemodelsfewshotlearners}
Tom~B. Brown, Benjamin Mann, Nick Ryder, Melanie Subbiah, Jared Kaplan, Prafulla Dhariwal, Arvind Neelakantan, Pranav Shyam, Girish Sastry, Amanda Askell, Sandhini Agarwal, Ariel Herbert-Voss, Gretchen Krueger, Tom Henighan, Rewon Child, Aditya Ramesh, Daniel~M. Ziegler, Jeffrey Wu, Clemens Winter, Christopher Hesse, Mark Chen, Eric Sigler, Mateusz Litwin, Scott Gray, Benjamin Chess, Jack Clark, Christopher Berner, Sam McCandlish, Alec Radford, Ilya Sutskever, and Dario Amodei. 2020.
\newblock \href {https://arxiv.org/abs/2005.14165} {Language models are few-shot learners}.
\newblock \emph{Preprint}, arXiv:2005.14165.

\bibitem[{Camerer et~al.(2003)Camerer, Ho, and Chong}]{camerer2003cognitive}
Colin Camerer, Teck Ho, and Juin-Kuan Chong. 2003.
\newblock A cognitive hierarchy theory of one-shot games and experimental analysis.
\newblock \emph{Available at SSRN 411061}.

\bibitem[{Chan et~al.(2024)Chan, Jiayang, Yim, Deng, Fan, Li, Liu, Zhang, Wang, and Song}]{chan-etal-2024-negotiationtom}
Chunkit Chan, Cheng Jiayang, Yauwai Yim, Zheye Deng, Wei Fan, Haoran Li, Xin Liu, Hongming Zhang, Weiqi Wang, and Yangqiu Song. 2024.
\newblock \href {https://doi.org/10.18653/v1/2024.findings-emnlp.244} {{N}egotiation{T}o{M}: A benchmark for stress-testing machine theory of mind on negotiation surrounding}.
\newblock In \emph{Findings of the Association for Computational Linguistics: EMNLP 2024}, pages 4211--4241, Miami, Florida, USA. Association for Computational Linguistics.

\bibitem[{Chen et~al.(2024)Chen, Wu, Zhou, Wen, Bi, Jiang, Cao, Hu, Lai, Xiong, and Huang}]{chen-etal-2024-tombench}
Zhuang Chen, Jincenzi Wu, Jinfeng Zhou, Bosi Wen, Guanqun Bi, Gongyao Jiang, Yaru Cao, Mengting Hu, Yunghwei Lai, Zexuan Xiong, and Minlie Huang. 2024.
\newblock \href {https://doi.org/10.18653/v1/2024.acl-long.847} {{T}o{MB}ench: Benchmarking theory of mind in large language models}.
\newblock In \emph{Proceedings of the 62nd Annual Meeting of the Association for Computational Linguistics (Volume 1: Long Papers)}, pages 15959--15983, Bangkok, Thailand. Association for Computational Linguistics.

\bibitem[{Cushman(2024)}]{cushman2024computational}
Fiery Cushman. 2024.
\newblock Computational social psychology.
\newblock \emph{Annual Review of Psychology}, 75(1):625--652.

\bibitem[{Devaine et~al.(2014)Devaine, Hollard, and Daunizeau}]{devaine2014theory}
Marie Devaine, Guillaume Hollard, and Jean Daunizeau. 2014.
\newblock Theory of mind: did evolution fool us?
\newblock \emph{PloS One}, 9(2):e87619.

\bibitem[{Gandhi et~al.(2023)Gandhi, Fränken, Gerstenberg, and Goodman}]{gandhi2023understandingsocialreasoninglanguage}
Kanishk Gandhi, Jan-Philipp Fränken, Tobias Gerstenberg, and Noah~D. Goodman. 2023.
\newblock \href {https://arxiv.org/abs/2306.15448} {Understanding social reasoning in language models with language models}.
\newblock \emph{Preprint}, arXiv:2306.15448.

\bibitem[{Gmytrasiewicz and Doshi(2005)}]{gmytrasiewicz_framework_2005}
Piotr~J Gmytrasiewicz and Prashant Doshi. 2005.
\newblock A framework for sequential planning in multi-agent settings.
\newblock \emph{Journal of Artificial Intelligence Research}, 24:49--79.

\bibitem[{Grant et~al.(2017)Grant, Nematzadeh, and Griffiths}]{grant2017can}
Erin Grant, Aida Nematzadeh, and Thomas~L Griffiths. 2017.
\newblock How can memory-augmented neural networks pass a false-belief task?
\newblock In \emph{CogSci}.

\bibitem[{Grosskopf and Nagel(2007)}]{grosskopf2007rational}
Brit Grosskopf and Rosemarie Nagel. 2007.
\newblock Rational reasoning or adaptive behavior? evidence from two-person beauty contest games.
\newblock \emph{Evidence from Two-Person Beauty Contest Games (June 2007). Harvard NOM Research Paper}, (01-09).

\bibitem[{Hou et~al.(2024)Hou, Zhang, Shen, Wu, and Lu}]{hou-etal-2024-timetom}
Guiyang Hou, Wenqi Zhang, Yongliang Shen, Linjuan Wu, and Weiming Lu. 2024.
\newblock \href {https://doi.org/10.18653/v1/2024.findings-acl.685} {{T}ime{T}o{M}: Temporal space is the key to unlocking the door of large language models{'} theory-of-mind}.
\newblock In \emph{Findings of the Association for Computational Linguistics: ACL 2024}, pages 11532--11547, Bangkok, Thailand. Association for Computational Linguistics.

\bibitem[{Hu and Shu(2023)}]{hu2023languagemodelsagentmodels}
Zhiting Hu and Tianmin Shu. 2023.
\newblock \href {https://arxiv.org/abs/2312.05230} {Language models, agent models, and world models: The law for machine reasoning and planning}.
\newblock \emph{Preprint}, arXiv:2312.05230.

\bibitem[{Huang et~al.(2024)Huang, La~Malfa, Marro, Asperti, Cohn, and Wooldridge}]{huang-etal-2024-notion}
X.~Angelo Huang, Emanuele La~Malfa, Samuele Marro, Andrea Asperti, Anthony~G. Cohn, and Michael~J. Wooldridge. 2024.
\newblock \href {https://doi.org/10.18653/v1/2024.findings-emnlp.167} {A notion of complexity for theory of mind via discrete world models}.
\newblock In \emph{Findings of the Association for Computational Linguistics: EMNLP 2024}, pages 2964--2983, Miami, Florida, USA. Association for Computational Linguistics.

\bibitem[{Jara-Ettinger(2019)}]{jara-ettinger_theory_2019}
Julian Jara-Ettinger. 2019.
\newblock Theory of mind as inverse reinforcement learning.
\newblock \emph{Current Opinion in Behavioral Sciences}, 29:105--110.

\bibitem[{Jin et~al.(2024)Jin, Wu, Cao, Xiang, Kuo, Hu, Ullman, Torralba, Tenenbaum, and Shu}]{jin-etal-2024-mmtom}
Chuanyang Jin, Yutong Wu, Jing Cao, Jiannan Xiang, Yen-Ling Kuo, Zhiting Hu, Tomer Ullman, Antonio Torralba, Joshua Tenenbaum, and Tianmin Shu. 2024.
\newblock \href {https://doi.org/10.18653/v1/2024.acl-long.851} {{MMT}o{M}-{QA}: Multimodal theory of mind question answering}.
\newblock In \emph{Proceedings of the 62nd Annual Meeting of the Association for Computational Linguistics (Volume 1: Long Papers)}, pages 16077--16102, Bangkok, Thailand. Association for Computational Linguistics.

\bibitem[{Jung et~al.(2024)Jung, Kim, Jin, Kim, Seonwoo, Choi, Oh, and Kim}]{jung-etal-2024-perceptions}
Chani Jung, Dongkwan Kim, Jiho Jin, Jiseon Kim, Yeon Seonwoo, Yejin Choi, Alice Oh, and Hyunwoo Kim. 2024.
\newblock \href {https://doi.org/10.18653/v1/2024.emnlp-main.1105} {Perceptions to beliefs: Exploring precursory inferences for theory of mind in large language models}.
\newblock In \emph{Proceedings of the 2024 Conference on Empirical Methods in Natural Language Processing}, pages 19794--19809, Miami, Florida, USA. Association for Computational Linguistics.

\bibitem[{Keysar et~al.(2003)Keysar, Lin, and Barr}]{keysar2003limits}
Boaz Keysar, Shuhong Lin, and Dale~J Barr. 2003.
\newblock Limits on theory of mind use in adults.
\newblock \emph{Cognition}, 89(1):25--41.

\bibitem[{Kim et~al.(2023)Kim, Sclar, Zhou, Bras, Kim, Choi, and Sap}]{kim-etal-2023-fantom}
Hyunwoo Kim, Melanie Sclar, Xuhui Zhou, Ronan Bras, Gunhee Kim, Yejin Choi, and Maarten Sap. 2023.
\newblock \href {https://doi.org/10.18653/v1/2023.emnlp-main.890} {{FANT}o{M}: A benchmark for stress-testing machine theory of mind in interactions}.
\newblock In \emph{Proceedings of the 2023 Conference on Empirical Methods in Natural Language Processing}, pages 14397--14413, Singapore. Association for Computational Linguistics.

\bibitem[{Kosinski(2023)}]{kosinski2023theory}
Michal Kosinski. 2023.
\newblock Theory of mind may have spontaneously emerged in large language models.
\newblock \emph{arXiv preprint arXiv:2302.02083}, 4:169.

\bibitem[{Kosinski(2024)}]{kosiniski_llm_tom_bench_mark_review}
Michal Kosinski. 2024.
\newblock \href {https://doi.org/10.1073/pnas.2405460121} {Evaluating large language models in theory of mind tasks}.
\newblock \emph{Proceedings of the National Academy of Sciences}, 121(45):e2405460121.

\bibitem[{Le et~al.(2019)Le, Boureau, and Nickel}]{le-etal-2019-revisiting}
Matthew Le, Y-Lan Boureau, and Maximilian Nickel. 2019.
\newblock \href {https://doi.org/10.18653/v1/D19-1598} {Revisiting the evaluation of theory of mind through question answering}.
\newblock In \emph{Proceedings of the 2019 Conference on Empirical Methods in Natural Language Processing and the 9th International Joint Conference on Natural Language Processing (EMNLP-IJCNLP)}, pages 5872--5877, Hong Kong, China. Association for Computational Linguistics.

\bibitem[{Leslie et~al.(2004)Leslie, Friedman, and German}]{leslie2004core}
Alan~M Leslie, Ori Friedman, and Tim~P German. 2004.
\newblock Core mechanisms in ‘theory of mind’.
\newblock \emph{Trends in cognitive sciences}, 8(12):528--533.

\bibitem[{Li et~al.(2023)Li, Chong, Stepputtis, Campbell, Hughes, Lewis, and Sycara}]{li-etal-2023-theory}
Huao Li, Yu~Chong, Simon Stepputtis, Joseph Campbell, Dana Hughes, Charles Lewis, and Katia Sycara. 2023.
\newblock \href {https://doi.org/10.18653/v1/2023.emnlp-main.13} {Theory of mind for multi-agent collaboration via large language models}.
\newblock In \emph{Proceedings of the 2023 Conference on Empirical Methods in Natural Language Processing}, pages 180--192, Singapore. Association for Computational Linguistics.

\bibitem[{Ma et~al.(2023{\natexlab{a}})Ma, Gao, and Xu}]{ma-etal-2023-tomchallenges}
Xiaomeng Ma, Lingyu Gao, and Qihui Xu. 2023{\natexlab{a}}.
\newblock \href {https://doi.org/10.18653/v1/2023.conll-1.2} {{T}o{MC}hallenges: A principle-guided dataset and diverse evaluation tasks for exploring theory of mind}.
\newblock In \emph{Proceedings of the 27th Conference on Computational Natural Language Learning (CoNLL)}, pages 15--26, Singapore. Association for Computational Linguistics.

\bibitem[{Ma et~al.(2023{\natexlab{b}})Ma, Sansom, Peng, and Chai}]{ma-etal-2023-towards-holistic}
Ziqiao Ma, Jacob Sansom, Run Peng, and Joyce Chai. 2023{\natexlab{b}}.
\newblock \href {https://doi.org/10.18653/v1/2023.findings-emnlp.72} {Towards a holistic landscape of situated theory of mind in large language models}.
\newblock In \emph{Findings of the Association for Computational Linguistics: EMNLP 2023}, pages 1011--1031, Singapore. Association for Computational Linguistics.

\bibitem[{Miura and Zilberstein(2021)}]{miura2021unifying}
Shuwa Miura and Shlomo Zilberstein. 2021.
\newblock A unifying framework for observer-aware planning and its complexity.
\newblock In \emph{Uncertainty in Artificial Intelligence}, pages 610--620. PMLR.

\bibitem[{Nematzadeh et~al.(2018)Nematzadeh, Burns, Grant, Gopnik, and Griffiths}]{nematzadeh-etal-2018-evaluating}
Aida Nematzadeh, Kaylee Burns, Erin Grant, Alison Gopnik, and Tom Griffiths. 2018.
\newblock \href {https://doi.org/10.18653/v1/D18-1261} {Evaluating theory of mind in question answering}.
\newblock In \emph{Proceedings of the 2018 Conference on Empirical Methods in Natural Language Processing}, pages 2392--2400, Brussels, Belgium. Association for Computational Linguistics.

\bibitem[{Ng and Russell(2000)}]{ng2000algorithms}
Andrew Ng and Stuart Russell. 2000.
\newblock Algorithms for inverse reinforcement learning.
\newblock \emph{ICML '00 Proceedings of the Seventeenth International Conference on Machine Learning}.

\bibitem[{Nickel et~al.(2024)Nickel, Schrewe, and Flek}]{nickel2024probingrobustnesstheorymind}
Christian Nickel, Laura Schrewe, and Lucie Flek. 2024.
\newblock \href {https://arxiv.org/abs/2410.06271} {Probing the robustness of theory of mind in large language models}.
\newblock \emph{Preprint}, arXiv:2410.06271.

\bibitem[{Pearl(2012)}]{pearl2012calculus}
Judea Pearl. 2012.
\newblock The do-calculus revisited.
\newblock \emph{arXiv preprint arXiv:1210.4852}.

\bibitem[{Pi et~al.(2024)Pi, Vadaparty, Bergen, and Jones}]{pi2024dissectingullmanvariationsscalpel}
Zhiqiang Pi, Annapurna Vadaparty, Benjamin~K. Bergen, and Cameron~R. Jones. 2024.
\newblock \href {https://arxiv.org/abs/2406.14737} {Dissecting the ullman variations with a scalpel: Why do llms fail at trivial alterations to the false belief task?}
\newblock \emph{Preprint}, arXiv:2406.14737.

\bibitem[{Premack and Woodruff(1978)}]{premack1978does}
David Premack and Guy Woodruff. 1978.
\newblock Does the chimpanzee have a theory of mind?
\newblock \emph{Behavioral and brain sciences}, 1(4):515--526.

\bibitem[{Qiu et~al.(2024)Qiu, Liu, Li, Zhu, and Zheng}]{qiu-etal-2024-minddial}
Shuwen Qiu, Mingdian Liu, Hengli Li, Song-Chun Zhu, and Zilong Zheng. 2024.
\newblock \href {https://doi.org/10.18653/v1/2024.sigdial-1.63} {{M}ind{D}ial: Enhancing conversational agents with theory-of-mind for common ground alignment and negotiation}.
\newblock In \emph{Proceedings of the 25th Annual Meeting of the Special Interest Group on Discourse and Dialogue}, pages 746--759, Kyoto, Japan. Association for Computational Linguistics.

\bibitem[{Rilling et~al.(2004)Rilling, Sanfey, Aronson, Nystrom, and Cohen}]{rilling2004neural}
James~K Rilling, Alan~G Sanfey, Jessica~A Aronson, Leigh~E Nystrom, and Jonathan~D Cohen. 2004.
\newblock The neural correlates of theory of mind within interpersonal interactions.
\newblock \emph{Neuroimage}, 22(4):1694--1703.

\bibitem[{Sap et~al.(2022)Sap, Le~Bras, Fried, and Choi}]{sap-etal-2022-neural}
Maarten Sap, Ronan Le~Bras, Daniel Fried, and Yejin Choi. 2022.
\newblock \href {https://doi.org/10.18653/v1/2022.emnlp-main.248} {Neural theory-of-mind? on the limits of social intelligence in large {LM}s}.
\newblock In \emph{Proceedings of the 2022 Conference on Empirical Methods in Natural Language Processing}, pages 3762--3780, Abu Dhabi, United Arab Emirates. Association for Computational Linguistics.

\bibitem[{Sarkadi et~al.(2019)Sarkadi, Panisson, Bordini, McBurney, Parsons, and Chapman}]{sarkadi2019modelling}
{\c{S}}tefan Sarkadi, Alison~R Panisson, Rafael~H Bordini, Peter McBurney, Simon Parsons, and Martin Chapman. 2019.
\newblock Modelling deception using theory of mind in multi-agent systems.
\newblock \emph{AI Communications}, 32(4):287--302.

\bibitem[{Schaafsma et~al.(2015)Schaafsma, Pfaff, Spunt, and Adolphs}]{schaafsma2015deconstructing}
Sara~M Schaafsma, Donald~W Pfaff, Robert~P Spunt, and Ralph Adolphs. 2015.
\newblock Deconstructing and reconstructing theory of mind.
\newblock \emph{Trends in cognitive sciences}, 19(2):65--72.

\bibitem[{Sclar et~al.(2023)Sclar, Kumar, West, Suhr, Choi, and Tsvetkov}]{sclar-etal-2023-minding}
Melanie Sclar, Sachin Kumar, Peter West, Alane Suhr, Yejin Choi, and Yulia Tsvetkov. 2023.
\newblock \href {https://doi.org/10.18653/v1/2023.acl-long.780} {Minding language models{'} (lack of) theory of mind: A plug-and-play multi-character belief tracker}.
\newblock In \emph{Proceedings of the 61st Annual Meeting of the Association for Computational Linguistics (Volume 1: Long Papers)}, pages 13960--13980, Toronto, Canada. Association for Computational Linguistics.

\bibitem[{Sclar et~al.(2022)Sclar, Neubig, and Bisk}]{sclar2022symmetric}
Melanie Sclar, Graham Neubig, and Yonatan Bisk. 2022.
\newblock Symmetric machine theory of mind.
\newblock In \emph{International Conference on Machine Learning}, pages 19450--19466. PMLR.

\bibitem[{Shapira et~al.(2024)Shapira, Levy, Alavi, Zhou, Choi, Goldberg, Sap, and Shwartz}]{shapira-etal-2024-clever}
Natalie Shapira, Mosh Levy, Seyed~Hossein Alavi, Xuhui Zhou, Yejin Choi, Yoav Goldberg, Maarten Sap, and Vered Shwartz. 2024.
\newblock \href {https://aclanthology.org/2024.eacl-long.138} {Clever hans or neural theory of mind? stress testing social reasoning in large language models}.
\newblock In \emph{Proceedings of the 18th Conference of the European Chapter of the Association for Computational Linguistics (Volume 1: Long Papers)}, pages 2257--2273, St. Julian{'}s, Malta. Association for Computational Linguistics.

\bibitem[{Sileo and Lernould(2023)}]{sileo-lernould-2023-mindgames}
Damien Sileo and Antoine Lernould. 2023.
\newblock \href {https://doi.org/10.18653/v1/2023.findings-emnlp.303} {{M}ind{G}ames: Targeting theory of mind in large language models with dynamic epistemic modal logic}.
\newblock In \emph{Findings of the Association for Computational Linguistics: EMNLP 2023}, pages 4570--4577, Singapore. Association for Computational Linguistics.

\bibitem[{Soubki et~al.(2024)Soubki, Murzaku, Yousefi~Jordehi, Zeng, Markowska, Mirroshandel, and Rambow}]{soubki-etal-2024-views}
Adil Soubki, John Murzaku, Arash Yousefi~Jordehi, Peter Zeng, Magdalena Markowska, Seyed~Abolghasem Mirroshandel, and Owen Rambow. 2024.
\newblock \href {https://doi.org/10.18653/v1/2024.findings-acl.880} {Views are my own, but also yours: Benchmarking theory of mind using common ground}.
\newblock In \emph{Findings of the Association for Computational Linguistics: ACL 2024}, pages 14815--14823, Bangkok, Thailand. Association for Computational Linguistics.

\bibitem[{Strachan et~al.(2024)Strachan, Albergo, Borghini, Pansardi, Scaliti, Gupta, Saxena, Rufo, Panzeri, Manzi et~al.}]{strachan2024testing}
James~WA Strachan, Dalila Albergo, Giulia Borghini, Oriana Pansardi, Eugenio Scaliti, Saurabh Gupta, Krati Saxena, Alessandro Rufo, Stefano Panzeri, Guido Manzi, et~al. 2024.
\newblock Testing theory of mind in large language models and humans.
\newblock \emph{Nature Human Behaviour}, 8(7):1285--1295.

\bibitem[{Sweller(1988)}]{sweller1988cognitive}
John Sweller. 1988.
\newblock Cognitive load during problem solving: Effects on learning.
\newblock \emph{Cognitive science}, 12(2):257--285.

\bibitem[{Tan et~al.(2024)Tan, Yeo, Jaidka, Wu, Xu, Jain, Chadha, Liu, and Ng}]{tan2024phantompersonabasedpromptingeffect}
Fiona~Anting Tan, Gerard~Christopher Yeo, Kokil Jaidka, Fanyou Wu, Weijie Xu, Vinija Jain, Aman Chadha, Yang Liu, and See-Kiong Ng. 2024.
\newblock \href {https://arxiv.org/abs/2403.02246} {Phantom: Persona-based prompting has an effect on theory-of-mind reasoning in large language models}.
\newblock \emph{Preprint}, arXiv:2403.02246.

\bibitem[{Ullman(2023)}]{ullman_large_2023}
Tomer Ullman. 2023.
\newblock \href {https://arxiv.org/abs/2302.08399} {Large language models fail on trivial alterations to theory-of-mind tasks}.
\newblock \emph{Preprint}, arXiv:2302.08399.

\bibitem[{Verbrugge and Mol(2008)}]{verbrugge2008learning}
Rineke Verbrugge and Lisette Mol. 2008.
\newblock Learning to apply theory of mind.
\newblock \emph{Journal of Logic, Language and Information}, 17:489--511.

\bibitem[{Wagner and Abend(2025)}]{wagner2025languagemodelprobabilitiesrepresent}
Eitan Wagner and Omri Abend. 2025.
\newblock \href {https://arxiv.org/abs/2505.02072} {What do language model probabilities represent? from distribution estimation to response prediction}.
\newblock \emph{Preprint}, arXiv:2505.02072.

\bibitem[{Wang et~al.(2025)Wang, Zhou, Sap, Forlizzi, and Shen}]{wang2025rethinkingtheorymindbenchmarks}
Qiaosi Wang, Xuhui Zhou, Maarten Sap, Jodi Forlizzi, and Hong Shen. 2025.
\newblock \href {https://arxiv.org/abs/2504.10839} {Rethinking theory of mind benchmarks for llms: Towards a user-centered perspective}.
\newblock \emph{Preprint}, arXiv:2504.10839.

\bibitem[{Wilf et~al.(2024)Wilf, Lee, Liang, and Morency}]{wilf-etal-2024-think}
Alex Wilf, Sihyun Lee, Paul~Pu Liang, and Louis-Philippe Morency. 2024.
\newblock \href {https://doi.org/10.18653/v1/2024.acl-long.451} {Think twice: Perspective-taking improves large language models{'} theory-of-mind capabilities}.
\newblock In \emph{Proceedings of the 62nd Annual Meeting of the Association for Computational Linguistics (Volume 1: Long Papers)}, pages 8292--8308, Bangkok, Thailand. Association for Computational Linguistics.

\bibitem[{Wimmer and Perner(1983)}]{wimmer1983beliefs}
Heinz Wimmer and Josef Perner. 1983.
\newblock Beliefs about beliefs: Representation and constraining function of wrong beliefs in young children's understanding of deception.
\newblock \emph{Cognition}, 13(1):103--128.

\bibitem[{Wu et~al.(2023)Wu, He, Jia, Mihalcea, Chen, and Deng}]{wu-etal-2023-hi}
Yufan Wu, Yinghui He, Yilin Jia, Rada Mihalcea, Yulong Chen, and Naihao Deng. 2023.
\newblock \href {https://doi.org/10.18653/v1/2023.findings-emnlp.717} {Hi-{T}o{M}: A benchmark for evaluating higher-order theory of mind reasoning in large language models}.
\newblock In \emph{Findings of the Association for Computational Linguistics: EMNLP 2023}, pages 10691--10706, Singapore. Association for Computational Linguistics.

\bibitem[{Xu et~al.(2024)Xu, Zhao, Zhu, Du, and He}]{xu-etal-2024-opentom}
Hainiu Xu, Runcong Zhao, Lixing Zhu, Jinhua Du, and Yulan He. 2024.
\newblock \href {https://doi.org/10.18653/v1/2024.acl-long.466} {{O}pen{T}o{M}: A comprehensive benchmark for evaluating theory-of-mind reasoning capabilities of large language models}.
\newblock In \emph{Proceedings of the 62nd Annual Meeting of the Association for Computational Linguistics (Volume 1: Long Papers)}, pages 8593--8623, Bangkok, Thailand. Association for Computational Linguistics.

\bibitem[{Ying et~al.(2025)Ying, Truong, Tenenbaum, and Gershman}]{ying_truong_tenenbaum_gershman_2025}
Lance Ying, Ryan Truong, Joshua Tenenbaum, and Samuel~J Gershman. 2025.
\newblock \href {osf.io/preprints/psyarxiv/4ctjp_v2} {Adaptive social learning using theory of mind}.

\bibitem[{Zhang et~al.(2023)Zhang, Du, Shan, Zhou, Du, Tenenbaum, Shu, and Gan}]{zhang2023building}
Hongxin Zhang, Weihua Du, Jiaming Shan, Qinhong Zhou, Yilun Du, Joshua~B Tenenbaum, Tianmin Shu, and Chuang Gan. 2023.
\newblock Building cooperative embodied agents modularly with large language models.
\newblock \emph{arXiv preprint arXiv:2307.02485}.

\bibitem[{Zhang and Gl{\"a}scher(2020)}]{zhang2020brain}
Lei Zhang and Jan Gl{\"a}scher. 2020.
\newblock A brain network supporting social influences in human decision-making.
\newblock \emph{Science advances}, 6(34):eabb4159.

\end{thebibliography}

\appendix

\section{A Case Study on Existing Benchmarks} \label{appendix:a}

Here we empirically demonstrate our arguments on two benchmarks, HiToM \cite{wu-etal-2023-hi} and FANToM \cite{kim-etal-2023-fantom}. These benchmarks were shown to be challenging even for strong models. 
We argue that the main challenge in these benchmarks is due to the complex reasoning required (type C errors) and not due to ToM. 

To test this, we design a simple test to predict the order of ToM in a question. In both benchmarks, samples are labeled with the order of ToM required to answer the question.
If models can determine the correct order of ToM but are incapable of answering the question, this indicates that the failure is not in ToM invocation.

\subsection{Experimental Setup}
\paragraph{Data.}

HiToM is a benchmark that is based on Sally-Anne-like stories. The stories include higher-order beliefs and questions about the stories range from factual questions (order $0$) to questions about $4$-th order beliefs. Each story has many questions in many orders.
Additionally, HiToM has stories with communication and deception, such that in many cases story characters intentionally report false statements.

FANToM consists of stories with conversations in which characters enter and leave. The questions are various. They include factual (order $0$) and belief questions, with belief questions including first- and second-order beliefs (the data includes the question type).

For evaluation, we use a set of $100$ random examples from each dataset. 
Some manual experiments were done, using a small number of examples. These examples were excluded from the evaluation.

\paragraph{Settings.}
We test whether the models can predict the ToM level of a question. We test once with the question alone and once with the story too. For each of these tests, we test with or without Chain of Thought (CoT) in the response.

\paragraph{Models.}
We test three models: GPT-3.5-turbo, GPT-4o-mini, and GPT-4o.\footnote{\url{https://platform.openai.com/docs/models/models}}

For HiToM, \citet{wu-etal-2023-hi} report an overall accuracy of $0.315$ with GPT-3.5-turbo and $0.59$ with GPT-4. For FANToM, an accuracy of $0.101$ is reported with GPT-3.5-turbo, $0.426$ with GPT-4-turbo, and $0.497$ with GPT-4-turbo.\footnote{\url{https://github.com/skywalker023/fantom}} These FANToM results are for the belief task with a short version of the story.

\begin{table*}
    \centering
    \begin{tabular}{c|c|cccc}
         \multirow{2}{*}{\textbf{Benchmark}} & \multirow{2}{*}{\textbf{Model}}  & \multicolumn{2}{c}{\textbf{Without Story}} & \multicolumn{2}{c}{\textbf{With Story}} \\
         & & \textbf{No-CoT } &\textbf{With-CoT} & \textbf{No-CoT } &\textbf{With-CoT}\\
         \hline \hline
        & GPT-3.5-turbo & 0.43 & 0.45 & 0.43 & 0.45\\
         HiToM & GPT-4o-mini & 0.81 & 0.81 & 0.81 & 0.81\\
         & GPT-4o & 0.92 & 1. & 0.92 & 1.\\
         \hline
         & GPT-3.5-turbo & 0.52 & 0.53 & 0.52 & 0.53\\
         FANToM& GPT-4o-mini & 0.84 & 0.84 & 0.84 & 0.84\\
         & GPT-4o & 0.86 & 0.92 & 0.86 & 0.92\\
    \end{tabular}
    \caption{ToM degree prediction results. We report the accuracy in multiple settings over $100$ random samples.}
    \label{tab:table1}
\end{table*}

\paragraph{Prompt.}
Here we describe the prompts used for the different settings.

For the setting with the story, we start with:
\begin{quote}
I will give you a story with a question.
\end{quote}

For the question alone, we start with:
\begin{quote}
I will give you a question about an unseen story.
\end{quote}

Then we give the explanation (in all cases):
\begin{quote}
I want you to tell me the Theory of Mind (ToM) order of the question. 
The ToM order is number of mentalization levels required to correctly answer the question. Where mentalization levels are the number of nested beliefs that one has to consider. 
For example, factual questions (e.g., "What day is it today?") require 0 levels of ToM, as they are invariant to beliefs. Questions about what someone else thinks (e.g., "What day does John think it is today?") require one level of ToM, as the fact itself is insufficient to answer the question.
Additional levels are possible, for example, ``What does John think that Mary thinks about what day it is today?'' requires two levels of mentalization.
    
Notice that I want the ToM order required to answer the specific question and not all mentalization processes that occur. So, for example, if the story is about John thinking about something and the question is whether John is thinking, the ToM order is 0. If the question is what John thinks, the ToM order is 1.
I also stress that for this task, we only address explicit beliefs and ignore implicit ones. So, for example, if the question is "Where does John take his dog on a walk?" the ToM order is 0, even though taking the dog involves an implicit mental process.
\end{quote}

In the no-CoT setting, we add the instruction:
\begin{quote}
Your answer must be in the format:
"The answer is: <answer>" 
Where <answer> is a number depicting the ToM order of the question.
Do NOT add any additional explanations.
\end{quote}

In the CoT setting, we add instead:
\begin{quote}
You can write explanations, but your last line must be in the format:
"The answer is: <answer>" 
Where <answer> is a number depicting the ToM order of the question.
\end{quote}

In the setting with the story, we input the story and the question:

\begin{quote}
The story is: 
<story>        

and the question is:
<question>
\end{quote}

Without the story, we input only the question:

\begin{quote}
The question is:
<question>
\end{quote}

\subsection{Results}

We report the results in Table \ref{tab:table1}. 

GPT-4o-mini achieves good results and GPT-4o is nearly perfect. GPT-3.5-turbo struggles with the task.
CoT gives some improvement. 

In all cases, the scores with and without the story as input are identical.

\subsection{Comments and Discussion}

In FANToM there are many cases of actions that include implicit ToM but are not considered as additional levels. 
For example, a question like ``What does [A] believe about how [B] overcomes the challenge of dealing with ...'' is marked as first-order since it only involves one explicit belief. However, it is also reasonable to classify this question as second-order since ``dealing'' implies some internal intention.

Some questions in FANToM are confusing as they contradict the answer. 
For example, as in the previous example, a question might ask what [A] believes about [B]'s actions, but according to the story [A] has no belief about it and this is the intended answer. This is confusing since the question suggests that [A] does have a belief.

In FANToM there is also no notion of deception -- all reports are assumed to be true. These assumptions are not even counted (in the dataset construction) as ToM. 
In a case like the previous example, if [B] said that he deals with the challenge in some way, a question about how he deals is considered zero-order. This consideration ignores an option that [B] was not telling the truth.

HiToM contains deception in some stories. However, it is always made clear to the reader what the truth is. It is also explicitly noted in the prompt that characters may not be reporting the truth.

We also note that in the HiToM paper, some analysis was done regarding the error types, with one type indicating \textit{insufficient reasoning-depth}. This is similar to our experiment, but it addresses explanations given by the model in the original task. It is possible that a model can predict the correct depth but get confused in the actual reasoning process leading to an incorrect stop. In our experiment, we test the depth independently of the question itself.

In conclusion, our experiment shows that the sub-task of predicting the ToM order is much easier than the original task. Even GPT-3.5-turbo, which struggles with this task, has better scores than on the original task. Additionally, this sub-task can be performed based on the question alone. 

The results indicate that the challenging part of these benchmarks is not the ToM order (=invoking ToM) but rather the inference given the order (Type C errors). Moreover, the benchmarks address the ToM level as a prediction task with a single correct solution. As we argue in the paper, this is a limited view of ToM.

\end{document}